\documentclass[runningheads]{llncs}

\usepackage[T1]{fontenc}                 
\usepackage{amsmath, amssymb, amsfonts} 
\usepackage{graphicx}                   
\usepackage{booktabs}                   
\usepackage{array}                      
\usepackage{multirow}                    
\usepackage[table]{xcolor}               
\usepackage{colortbl}
\usepackage[most]{tcolorbox}            
\usepackage{float}                       
\usepackage{hyperref}                   
\usepackage{textcomp}                    
\usepackage{algorithmic}                 
\usepackage{placeins}                    
\usepackage{cite}                       
\usepackage{tikz}                        %

\usetikzlibrary{shapes.geometric, arrows, positioning, calc, decorations.pathreplacing, arrows.meta, fit} 

\def\BibTeX{{\rm B\kern-.05em{\sc i\kern-.025em b}\kern-.08em
    T\kern-.1667em\lower.7ex\hbox{E}\kern-.125emX}}

\tikzstyle{box} = [rectangle, rounded corners, minimum width=3cm, minimum height=1cm, text centered, draw=black, fill=yellow!20]
\tikzstyle{innerbox} = [rectangle, rounded corners, minimum width=3cm, minimum height=0.8cm, text centered, draw=black, fill=blue!10]
\tikzstyle{connector} = [->, thick]
\tikzstyle{dashedconnector} = [->, dashed, thick]
\tikzstyle{labelbox} = [rectangle, minimum width=2cm, minimum height=0.8cm, text centered, draw=black, fill=green!20]

\begin{document}

\title{Few-Shot Optimized Framework for Hallucination Detection in Resource-Limited NLP Systems
}

\author{
    Baraa Hikal\inst{1}\and
    Ahmed Nasreldin\inst{1}\and
    Ali Hamdi\inst{1} \and
    Ammar Mohammed\inst{1}\
}

\institute{
    Faculty of Computer Science, MSA University, Giza, Egypt\\
    \email{\{baraa.moaweya, ahmed.nasreldin, ahamdi, ammar.mohammed\}@msa.edu.eg}
}
\authorrunning{B. Hikal}

\maketitle

\begin{abstract}
Hallucination detection in text generation remains an ongoing struggle for natural language processing (NLP) systems, frequently resulting in unreliable outputs in applications such as machine translation and definition modeling. Existing methods struggle with data scarcity and the limitations of unlabeled datasets, as highlighted by the SHROOM shared task at SemEval-2024. In this work, we propose a novel framework to address these challenges, introducing \textbf{DeepSeek Few-shot Optimization} to enhance weak label generation through iterative prompt engineering. We achieved high-quality annotations that considerably enhanced the performance of downstream models by restructuring data to align with instruct generative models. We further fine-tuned the \textbf{Mistral-7B-Instruct-v0.3} model on these optimized annotations, enabling it to accurately detect hallucinations in resource-limited settings. Combining this fine-tuned model with ensemble learning strategies, our approach achieved \textbf{85.5\% accuracy} on the test set, setting a new benchmark for the SHROOM task. This study demonstrates the effectiveness of data restructuring, few-shot optimization, and fine-tuning in building scalable and robust hallucination detection frameworks for resource-constrained NLP systems.
\end{abstract}

\keywords{few shot, transformer architectures, hallucination detection, generative NLP, weak labeling, prompt engineering}

\section{Introduction}

Hallucination detection in text generation has emerged as one of the most pressing challenges in natural language processing (NLP) \cite{OctavianB2024,Muller2023domain}. Despite the remarkable advancements made in AI systems, hallucinations—outputs that are fluent yet factually incorrect—continue to compromise the reliability of these technologies \cite{Dale2023detecting}. This issue is especially problematic in applications such as machine translation and definition modeling, where factual accuracy and credibility are essential. Left unaddressed, hallucinations not only limit the practical usability of NLP systems but also pose risks in high-stakes domains like healthcare, legal documentation, and scientific research \cite{guerreiro2023looking}.

Existing methods for hallucination detection can be broadly categorized into two main approaches: \textit{model-aware} and \textit{model-agnostic}. Model-aware techniques leverage access to a model's internal workings, such as weights and logits, while model-agnostic methods, often referred to as black-box approaches, rely solely on the generated text without insights into the model's internals \cite{liu2024hit, Dale2023detecting}. Within these categories, supervised learning is a frequently used technique that uses a large quantity of labeled data to train models on annotated datasets in order to detect hallucinations, which can be challenging to acquire \cite{arzt2024tusemevalSLPL}.Weak supervision overcomes this constraint by labeling unlabeled training data with noisy or heuristically generated labels, such as those produced by large language models (LLMs) like GPT-4 \cite{OPDAI2024}. While this allows for scalability beyond manually annotated datasets, it often struggles with noisy labels that may fail to capture the complexities of hallucination detection \cite{mehta2024halu}. Few-shot learning presents an alternative approach by training models on a limited number of labeled examples, requiring precise optimization of prompts and instructions to ensure reliable performance \cite{guerreiro2023looking}. In addition to these core methods, complementary techniques such as prompt engineering and data restructuring have remarkably improved hallucination detection. Prompt engineering entails formulating specific prompts to guide LLMs in detecting hallucinations, improving accuracy and reliability \cite{OctavianB2024}. Data restructuring, on the other hand, reorganizes and refines datasets to align with the capabilities of modern generative models, involving techniques such as selecting relevant data points, filtering out extraneous information, and adapting data formats for particular models \cite{Muller2023domain}. These methods collectively address both the fluency and factual accuracy of generated text but face never-ending challenges, particularly in resource-constrained scenarios where labeled data is scarce \cite{fallah2024slplsemeval}. Current research aims to develop frameworks that effectively utilize limited data while addressing the inherent complexities of hallucination detection tasks \cite{Rykov2024smurfcat}.

This paper presents a framework that leverages data restructuring and weak label optimization to address these limitations. By integrating few-shot prompt engineering with model fine-tuning, we aim to bridge the gap between data scarcity and the demand for robust detection systems. Our approach demonstrates that combining structured data processing with ensemble learning strategies can deliver scalable and accurate solutions for detecting hallucinations in resource-limited NLP applications.

\begin{figure}[H]
\centering
\label{fig:workflow_diagram}
\scalebox{0.6}{ 
\definecolor{peach}{RGB}{250, 229, 225}
\definecolor{terq}{RGB}{57, 171, 189}
\definecolor{headerblue}{HTML}{DAE8FC}  
\definecolor{nodegreen}{HTML}{D5E8D4}  
\definecolor{nodeyellow}{HTML}{FFF2CC} 
\definecolor{arrowcolor}{HTML}{5F9EA0} 
\begin{tikzpicture}[
    node distance=1cm and 1.5cm, 
    font=\rmfamily\small,
    box/.style={rectangle, draw, rounded corners, thick, text centered, minimum width=3cm, minimum height=0.8cm, fill=white!90, draw=black}, 
    subbox/.style={rectangle, draw, thick, rounded corners, fill=white!90, text centered, minimum width=2.7cm, minimum height=0.6cm, draw=black}, 
    flowarrow/.style={thick, ->, >=stealth, draw=black}, 
    dashedarrow/.style={thick, dashed, ->, >=stealth, draw=black, line width=0.4mm} 
]
    
\node[box, fill=red!30] (dataset) {\textbf{Unlabeled Training Data}};

\node[rectangle, draw, thick, rounded corners, fill=gray!10, minimum width=5cm, minimum height=2.5cm, right=2cm of dataset] (prompt_module) {};
\node[font=\bfseries\small] at ([yshift=-0.3cm]prompt_module.north) {DeepSeek-v3};
\node[subbox, draw=terq, below=0.6cm of prompt_module.north, fill=peach, line width=0.4mm] (prompt_design) {\textbf{System Instructions}};
\node[subbox, below=0.3cm of prompt_design, fill=peach, draw=terq, line width=0.4mm] (few_shot) {\textbf{Few-Shot Prompting}};

\node[box, fill=headerblue, right=3cm of prompt_module, yshift=-0.1cm] (weak_set) {\textbf{Weakly Supervised Data}};

\coordinate (midpoint) at ($(prompt_design.east)!0.5!(few_shot.east) + (2.2cm, 0)$);

\node[box, fill=nodeyellow, below=1.5cm of weak_set, yshift=-0.3cm] (data_prep) {\textbf{Data Restructure}};

\node[rectangle, draw, thick, rounded corners, fill=gray!10, minimum width=5cm, minimum height=1.5cm, left=3.6cm of data_prep] (fine_tuning) {};
\node[font=\bfseries\small] at ([yshift=-0.3cm]fine_tuning.north) {Model Fine-Tuning};
\node[subbox, below=0.65cm of fine_tuning.north, fill=peach, draw=terq, line width=0.4mm] (base_model) {\textbf{Mistral-7B-Instruct-v0.3}};

\node[box, fill=nodegreen!60, left=2.1cm of fine_tuning] (halluc_detector) {\textbf{Hallucination Detector}};

\draw[flowarrow] (dataset) -- (prompt_module);

\draw[dashedarrow, -] (prompt_design.east) to[out=0, in=180] (midpoint);
\draw[dashedarrow, -] (few_shot.east) to[out=0, in=180] (midpoint);
\draw[dashedarrow] (midpoint) -- (weak_set.west);

\draw[flowarrow] (weak_set.south) -- (data_prep.north);

\draw[flowarrow] (data_prep.west) -- (fine_tuning.east);

\draw[flowarrow] (fine_tuning.west) -- (halluc_detector.east);

\end{tikzpicture}
}
\caption{Workflow diagram for weakly supervised data generation and fine-tuning.}

\end{figure}
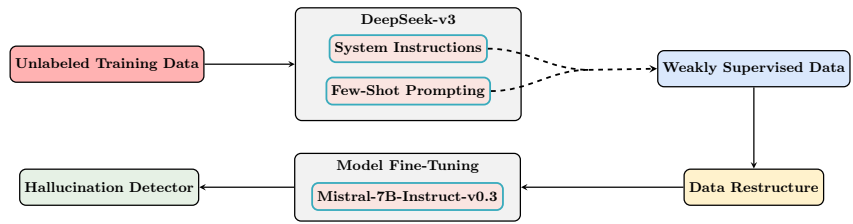

\section{Related Work}

Detection of hallucinations in NLG is an important area for research that has been considered as a key for increasing the reliability of systems in machine translation, summarization, and definition modeling. The recent works that have been carried out include NLI, semantic similarity measures, fine-tuning of pre-trained models, LLMs, and prompt engineering \cite{mehta2024halu, mickus2024semeval}. These methods, while having achieved certain successes, often suffer from problems of data scarcity, noisy labels, and generalization issues; hence, our framework overcomes those using DeepSeek Few-shot Optimization and task-specific fine-tuning.

Natural language inference detects inconsistencies by evaluating semantic relationships between source and output, particularly in summarization tasks \cite{borra2024malto}. However, it requires large annotated datasets, which are difficult to obtain. Our framework uses weak supervision and few-shot learning to generate high-quality labels with minimal annotated data. By iteratively refining prompts and system instructions, we address the limitations of traditional NLI methods.

Semantic similarity measures like ROUGE and BLEU are limited by their reliance on lexical overlap \cite{mickus2024semeval}, leading to embedding-based techniques like cosine similarity with SentenceTransformers. While these methods improve over traditional metrics, they still struggle with noisy labels. Our approach restructures datasets into a generative format, enabling models like Mistral-7B-Instruct-v0.3 to leverage their strengths in natural language understanding and generation.

Fine-tuning pre-trained models like BERT, DeBERTa, and BART on labeled datasets has shown promise for domain-specific hallucination detection \cite{MARiA2024}. However, reliance on annotated data limits generalization. LLMs have emerged as a versatile alternative \cite{pan2024umuteam}, with approaches like Self-CheckGPT analyzing response consistency \cite{manakul2023selfcheckgpt} and TrueTeacher generating synthetic training data \cite{liu2023mitigating}. Our work builds on these advancements by introducing DeepSeek Few-shot Optimization, which iteratively refines weak labels through task-specific system instructions and few-shot prompting.

Prompt engineering has proven effective in scenarios with limited labeled data, using techniques like few-shot learning and Chain-of-Thought prompting \cite{liu2023mitigating}. However, these methods often rely on generic prompts. Our framework integrates task-specific system instructions and 8-shot prompting, enabling DeepSeek-v3 to generate high-quality weak labels even with minimal examples.

Benchmark datasets like HaluEval have been critical for evaluating hallucination detection methods \cite{mickus2024semeval,lei2023chain}, but the scarcity of multilingual datasets limits generalization. The SemEval-2024 SHROOM shared task has advanced weak supervision, prompt engineering, and ensemble learning \cite{griogoriadou2024ails}. Our work builds on these advancements, demonstrating the effectiveness of our approach in resource-constrained settings.

While we've made significant strides, there are still challenges to overcome in making models more interpretable, expanding their ability to handle multiple languages, and detecting when the model generates incorrect information at the token level. Our research tackles these issues by using few-shot optimization and ensemble learning, which makes the models more robust and scalable. In the future, we could explore ways to incorporate features that are specific to the model and improve how well the model can generalize across different tasks.
\section{Methodology}

\subsection{Prompt Engineering for Enhancing Weak Label Generation}

When training models to detect hallucinations, the quality of weak labels plays a key role, especially when labeled data is limited. In this section, we walk through how we refined prompt engineering techniques with \textbf{DeepSeek-v3} to create enhanced weak labels. We provide a detailed explanation of the mathematical modeling and iterative processes involved.

\subsubsection{Initial Configuration}

The process begins with the deployment of \textbf{DeepSeek-v3} in its default configuration to generate weak labels for the training dataset. The model's output probability for label \( y \) given input \( x \), default prompt \( p_{\text{default}} \), and generic system instruction \( s_{\text{generic}} \) is computed as:

\[
P_{\text{initial}}(y|x) = \text{softmax}\left(\mathbf{W}_{\text{initial}} \cdot f(x, p_{\text{default}}, s_{\text{generic}}) + \mathbf{b}_{\text{initial}}\right),
\]

where:
\begin{itemize}
    \item \( f(x, p, s) \) is a function that combines the embeddings of the input \( x \), prompt \( p \), and system instructions \( s \);
    \item \( \mathbf{W}_{\text{initial}} \) is the weight matrix for the initial configuration;
    \item \( \mathbf{b}_{\text{initial}} \) is the bias vector, which introduces a shift in the linear transformation to improve model flexibility.
\end{itemize}

This configuration achieves a baseline accuracy of \textbf{73.6\%}.

\subsubsection{Refinement with Task-Specific System Instructions}

To improve the quality of weak labels, task-specific system instructions \( s_{\text{task}} \) are introduced. The refined output probability is given by:

\[
P_{\text{refined}}(y|x) = \text{softmax}\left(\mathbf{W}_{\text{refined}} \cdot f(x, p_{\text{default}}, s_{\text{task}}) + \mathbf{b}_{\text{refined}}\right),
\]

where:
\begin{itemize}
    \item \( \mathbf{W}_{\text{refined}} \) is the weight matrix after incorporating task-specific instructions;
    \item \( \mathbf{b}_{\text{refined}} \) is the refined bias vector.
\end{itemize}

This refinement aligns the model's output more closely with the hallucination detection task, increasing accuracy to \textbf{77.1\%}.

\subsubsection{Few-Shot Prompting with System Instructions}

Further improvements are achieved by leveraging an 8-shot prompting strategy. The prompt \( p_{\text{8shot}} \) includes eight examples of accurately labeled data, and the output probability is computed as:

\[
P_{\text{fewshot}}(y|x) = \text{softmax}\left(\mathbf{W}_{\text{fewshot}} \cdot f(x, p_{\text{fewshot}}, s_{\text{task}}) + \mathbf{b}_{\text{fewshot}}\right),
\]

where:
\begin{itemize}
    \item \( \mathbf{W}_{\text{fewshot}} \) is the weight matrix after incorporating few-shot prompting;
    \item \( \mathbf{b}_{\text{fewshot}} \) is the bias vector for the few-shot configuration.
\end{itemize}

This approach enhances the model's ability to generate high-quality weak labels, achieving an accuracy of \textbf{82.4\%} on the validation set.

\subsubsection{Iterative Refinement Process}

The iterative refinement process involves optimizing the system instructions \( s_{\text{task}} \) to maximize an evaluation metric. This can be formalized as:

\[
s_{\text{task}}^* = \arg\max_{s} \text{Eval}\left(P(y|x, p_{\text{fewshot}}, s), y_{\text{true}}\right),
\]

where \( \text{Eval} \) is a function that measures the alignment between the predicted probabilities \( P(y|x, p, s) \) and the true labels \( y_{\text{true}} \). This iterative process ensures continuous improvement in label quality.

\subsection{Dataset Labeling and Reconstruction}

Using the optimized configuration, the unlabeled training data was processed to generate weak labels tailored to the hallucination detection task. These labels were subsequently used to prepare a dataset for fine-tuning the \textbf{Mistral-7B-Instruct-v0.3} model.

To maximize the fine-tuning potential of the generated dataset, the raw classification data was restructured into a format suitable for generative modeling.
The reconstruction process reformulated the task, allowing the generative model to leverage its strengths in both natural language understanding and generation.

Figure~\ref{fig:before_reconstruction} shows the original dataset, and Table~\ref{tab:reconstructed_dataset_example} its generative conversion.

\begin{figure}[H]
\centering
\caption{Example of the dataset in its original format, intended for classification tasks.}
\label{fig:before_reconstruction}
\begin{tcolorbox}[colback=gray!10, colframe=gray!50, sharp corners, boxrule=0.4mm, breakable, width=0.8\textwidth]
\small
\texttt{
\ \\
\ \ "hyp": "Resembling or characteristic of a weasel.", \\
\ \ "ref": "tgt", \\
\ \ 
\ \ "tgt": "Resembling a weasel (in appearance).", \\
\ \ "task": "DM", \\
\ \ "label": "Not Hallucination", \\
\
}
\end{tcolorbox}

\end{figure}

{
\setlength{\tabcolsep}{12pt} 
\renewcommand{\arraystretch}{1.8} 
\begin{table}[H]
\centering
\caption{Example of Reconstructed Dataset for Generative Models}
\label{tab:reconstructed_dataset_example}
\setlength{\tabcolsep}{12pt} 
\renewcommand{\arraystretch}{1.8} 
\begin{tabular}{|>{\bfseries}l|p{9cm}|}
\hline
\rowcolor[HTML]{E0E0E0} 
\textbf{Role} & \textbf{Content} \\ \hline
System & 
You are a model that decides if the Sentence is Hallucination or Not Hallucination. \\ \hline
\rowcolor[HTML]{F7F7F7} 
User & 
Context: Resembling a weasel (in appearance). Sentence: Resembling or characteristic of a weasel. Is the Sentence hallucinated or not? \\ \hline
Assistant & 
Not Hallucination \\ \hline
\end{tabular}
\end{table}

}

\subsection{Model Fine-Tuning}

The weakly labeled dataset, consisting of 30,000 samples, served as the foundation for fine-tuning the \textbf{Mistral-7B-Instruct-v0.3} model. Fine-tuning was conducted with a carefully designed configuration to ensure alignment with the nuances of hallucination detection. The key parameters of the training configuration are summarized in Table~\ref{tab:training_configuration}.

\begin{table}[H]
\centering
\caption{Training Configuration for Fine-Tuning Mistral-7B-Instruct-v0.3}
\label{tab:training_configuration}
\begin{tabular}{ll}
\toprule
\textbf{Parameter} & \textbf{Value} \\
\midrule
Batch Size & 8 \\
Learning Rate & 2e-5 \\
Training Steps & 500 \\
Optimizer & AdamW \\
Fine-Tuning & LoRA (Low-Rank Adaptation) \\
LoRA Rank & 64 \\
Hardware & NVIDIA L40S GPUs \\
\bottomrule
\end{tabular}
\end{table}

This process adapted the model to the specific requirements of hallucination detection, enabling it to effectively utilize the weakly labeled data for improved performance in downstream tasks.

\section{Results and Discussion}
\subsection{Few-Shot Optimization for Weak Label Generation}
A cornerstone of this study is the implementation of few-shot optimization for generating high-quality weak labels using \textbf{DeepSeek-v3}. By employing an 8-shot prompting strategy combined with task-specific system instructions, the weak label generation process effectively bridged the gap between data scarcity and the demand for high-quality training data. This approach ensured that even with limited annotated resources, the weak labels aligned closely with task requirements, providing a robust foundation for fine-tuning. 

Our findings show just how important few-shot optimization can be when resources are constrained. By helping DeepSeek learn effectively from just a handful of examples, this approach not only improved the quality of weak supervision but also minimized the need for massive amounts of labeled data. This makes few-shot learning a practical and scalable solution for tackling tough problems like hallucination detection and other NLP challenges.

\begin{table}[H]
\centering
\caption{Prompt Engineering Approaches and Corresponding Accuracies on Validation Dataset.}
\label{tab:prompt_engineering}
\begin{tabular}{lccc}
\toprule
\textbf{Approach} & \textbf{Accuracy (\%)} \\
\midrule
DeepSeek-v3 (Default) & 73.6 \\
DeepSeek-v3 with System Instructions & 77.1 \\
DeepSeek-v3 with 8-Shot + System Instructions & 82.4 \\
\bottomrule
\end{tabular}
\end{table}

\subsection{Ensemble Learning for Enhanced Robustness}
To maximize performance, we aggregated predictions from seven independently fine-tuned checkpoints of \textbf{Mistral-7B-Instruct-v0.3} using majority voting. This ensemble strategy leveraged the diversity of individual models, effectively reducing biases and improving overall robustness. The ensemble achieved a test accuracy of \textbf{85.5\%}, establishing a new benchmark for hallucination detection in the SHROOM shared task.

The effectiveness of ensemble learning underscores its role as a complementary strategy to few-shot optimization and fine-tuning. By combining the strengths of multiple models, the ensemble approach provides enhanced generalizability and reliability across diverse test cases.

\begin{table}[H]
\centering
\caption{Model Voting Results for the Test Set on the Model-Agnostic Track}
\label{tab:model_voting_results}
\begin{tabular}{lr}
\toprule
\textbf{Model Variant} & \textbf{Accuracy} \\
\midrule
Mistral-7B-Instruct-v0.3-v0 & 0.838 \\
Mistral-7B-Instruct-v0.3-v1 & 0.844 \\
Mistral-7B-Instruct-v0.3-v2 & 0.841 \\
Mistral-7B-Instruct-v0.3-v3 & 0.845 \\
Mistral-7B-Instruct-v0.3-v5 & 0.832 \\
Mistral-7B-Instruct-v0.3-v6 & 0.834 \\
Mistral-7B-Instruct-v0.3-v8 & 0.845 \\
\textbf{Ensemble Result} & \textbf{0.855} \\
\bottomrule
\end{tabular}
\end{table}

\subsection{Model-Agnostic Track Rankings}
The proposed system achieved a top ranking on the model-agnostic track of the SHROOM shared task, with an accuracy of \textbf{0.855}. Table~\ref{tab:model_agnostic_results} summarizes the rankings, highlighting the competitive performance of the framework. Our findings highlight how powerful few-shot optimization and ensemble learning can be, showing they can achieve top-tier results in detecting hallucinations.

\begin{table}[H]
\centering
\caption{Model-Agnostic Track Rankings Based on Accuracy (Acc)}
\label{tab:model_agnostic_results}
\rowcolors{2}{gray!15}{white}
\begin{tabular}{llc}
\toprule
\textbf{Rank} & \textbf{System} & \textbf{Accuracy (Acc)} \\
\midrule
\textbf{1} & \textbf{Ours} & \textbf{0.855} \\
2 & Halu-NLP \cite{mehta2024halu} & 0.847 \\
3 & OPDAI \cite{OPDAI2024} & 0.836 \\
4 & HIT-MI\&T Lab \cite{liu2024hit} & 0.831 \\
5 & SHROOM-INDELab \cite{allen2024shroom} & 0.829 \\
6 & Alejandro Mosquera & 0.826 \\
7 & DeepPavlov & 0.821 \\
8 & BruceW & 0.821 \\
9 & TU Wien \cite{arzt2024tusemevalSLPL} & 0.817 \\
10 & SmurfCat \cite{Rykov2024smurfcat} & 0.814 \\
\bottomrule
\end{tabular}
\end{table}

\subsection{Discussion}
The results underscore the transformative potential of few-shot optimization for hallucination detection. By leveraging minimal labeled examples, the framework effectively addressed data scarcity challenges, producing weak labels that were highly aligned with task-specific requirements. This approach not only highlights the scalability of few-shot learning but also demonstrates its utility in enhancing weak supervision for complex NLP tasks.

The ensemble learning strategy further amplified the framework’s robustness, achieving a benchmark-setting accuracy of \textbf{85.5\%}. These findings validate the synergy between few-shot optimization, task-specific fine-tuning, and ensemble learning in tackling hallucination detection. Future work could explore extending this framework to other NLP tasks, incorporating model-aware features, or refining cross-task generalization to further improve performance and adaptability.

\section{Conclusion}

In this study, we tackle the challenge of detecting hallucinations in text generation systems by developing a strong approach that works even when labeled data is limited. Leveraging \textbf{DeepSeek-v3 \cite{DeepSeek}} for pseudo-label generation through 8-shot prompt engineering and task-specific instructions, we created a high-quality weakly supervised dataset that significantly improved model performance.

To optimize fine-tuning, we have restructured the dataset into a generative text style aligned with the capabilities of the \textbf{Mistral-7B-Instruct-v0.3} model. Using \textbf{LoRA} (Low-Rank Adaptation), we efficiently fine-tuned the model, achieving \textbf{85.5\% accuracy} on the model-agnostic track of the SHROOM shared task. By combining multiple fine-tuned models, ensemble learning pushed accuracy even higher, showing how powerful this approach can be.

This work points to the importance of data alignment with model architectures and also leveraging weak supervision to improve the state-of-the-art performance for NLP tasks. Future work will extend this framework on model-aware tracks and explore hybrid ensemble techniques that can further improve robustness and accuracy.


\begin{thebibliography}{10}
\providecommand{\url}[1]{\texttt{#1}}
\providecommand{\urlprefix}{URL }
\providecommand{\doi}[1]{https://doi.org/#1}

\bibitem{DeepSeek}
AI@DeepSeek: Deepseek-v3: A versatile and high-performance language model. arXiv preprint arXiv:2412.19437  (2024), \url{https://arxiv.org/abs/2412.19437}

\bibitem{allen2024shroom}
Allen, B.P., Polat, F., Groth, P.: {SHROOM}-{INDElab} at {SemEval}-2024 task 6: Zero- and few-shot {LLM}-based classification for hallucination detection. In: Proceedings of the 18th International Workshop on Semantic Evaluation (SemEval-2024). Association for Computational Linguistics (2024)

\bibitem{arzt2024tusemevalSLPL}
Arzt, V., Azarbeik, M.M., Lasy, I., Kerl, T., Recski, G.: {TU} wien at {SemEval-2024} task 6: Unifying model-agnostic and model-aware techniques for hallucination detection. In: Proceedings of the 18th International Workshop on Semantic Evaluation (SemEval-2024). pp. 1172--1186. Association for Computational Linguistics (2024)

\bibitem{borra2024malto}
Borra, F., Savelli, C., Rosso, G., Koudounas, A., Giobergia, F.: {MALTO} at {SemEval}-2024 task 6: Leveraging synthetic data for {LLM} hallucination detection. In: Proceedings of the 18th International Workshop on Semantic Evaluation (SemEval-2024). pp. 1793--1808. Association for Computational Linguistics (2024)

\bibitem{OctavianB2024}
Brodoceanu, O.: Octavianb at semeval-2024 task 6: An exploration of humanlike qualities in hallucinated llm texts. In: Proceedings of the 18th International Workshop on Semantic Evaluation (SemEval-2024). pp. 1162--1165 (2024), \url{https://aclanthology.org/2024.semeval-1.169/}

\bibitem{Dale2023detecting}
Dale, D., Voita, E., Barrault, L., Costa-juss{\`a}, M.R.: Detecting and mitigating hallucinations in machine translation: Model internal workings alone do well, sentence similarity even better. In: Proceedings of the 61st Annual Meeting of the Association for Computational Linguistics. vol.~1, pp. 36--50. Association for Computational Linguistics, Toronto, Canada (July 9-14 2023), \url{https://aclanthology.org/2023.acl-long.36/}

\bibitem{fallah2024slplsemeval}
Fallah, P., Gooran, S., Jafarinasab, M., Sadeghi, P., Farnia, R., Tarabkhah, A., Taghavi, Z.S., Sameti, H.: {SLPL} {SHROOM} at {SemEval2024} task 06: A comprehensive study on models ability to detect hallucination. In: Proceedings of the 18th International Workshop on Semantic Evaluation (SemEval-2024). pp. 1137--1143. Association for Computational Linguistics (2024)

\bibitem{griogoriadou2024ails}
Griogoriadou, N., Lymperaiou, M., Filandrianos, G., Stamou, G.: Ails-ntua at semeval-2024 task 6: Efficient model tuning for hallucination detection and analysis. In: Proceedings of the 18th International Workshop on Semantic Evaluation (SemEval-2024). pp. 1551--1557 (2024)

\bibitem{guerreiro2023looking}
Guerreiro, N.M., Voita, E., Martins, A.: Looking for a needle in a haystack: A comprehensive study of hallucinations in neural machine translation. In: Proceedings of the 17th Conference of the European Chapter of the Association for Computational Linguistics. pp. 1059--1075. Association for Computational Linguistics (2023)

\bibitem{lei2023chain}
Lei, D., Li, Y., Hu, M.M., Wang, M., Yun, V., Ching, E., Kamal, E.: Chain of natural language inference for reducing large language model ungrounded hallucinations. In: Proceedings of the 2023 Conference on Empirical Methods in Natural Language Processing. pp. 3445--3454 (2023)

\bibitem{liu2023mitigating}
Liu, F., Lin, K., Li, L., Wang, J., Yacoob, Y., Wang, L.: {MITIGATING HALLUCINATION IN LARGE MULTIMODAL MODELS VIA ROBUST INSTRUCTION TUNING}. arXiv preprint arXiv:2311.05556  (2023)

\bibitem{liu2024hit}
Liu, W., Shi, W., Zhang, Z., Huang, H.: {HIT}-{MI\&T} lab at {SemEval}-2024 task 6: {DeBERTa}-based entailment model is a reliable hallucination detector. In: Proceedings of the 18th International Workshop on Semantic Evaluation (SemEval-2024). Association for Computational Linguistics (2024)

\bibitem{manakul2023selfcheckgpt}
Manakul, P., Goyal, M., Chowdhury, A., Raghunathan, A., Bhardwaj, A.: Selfcheckgpt: Zero-resource detection of machine-generated text. arXiv preprint arXiv:2303.08896  (2023)

\bibitem{mehta2024halu}
Mehta, R., Hoblitzell, A., O’Keefe, J., Jang, H., Varma, V.: {Halu}-{NLP} at {SemEval}-2024 task 6: {MetaCheckGPT} - a multi-task hallucination detection using {LLM} uncertainty and meta-models. In: Proceedings of the 18th International Workshop on Semantic Evaluation (SemEval-2024). Association for Computational Linguistics (2024)

\bibitem{mickus2024semeval}
Mickus, T., Zosa, E., V{\'a}zquez, R., Vahtola, T., Tiedemann, J., Segonne, V., Raganato, A., Apidianaki, M.: {SemEval}-2024 shared task 6: {SHROOM}, a shared-task on hallucinations and related observable overgeneration mistakes. In: Proceedings of the 18th International Workshop on Semantic Evaluation (SemEval-2024). pp. 1980--1994. Association for Computational Linguistics (2024)

\bibitem{Muller2023domain}
M{\"u}ller, M., Rios, A., Sennrich, R.: Domain robustness in neural machine translation. In: Proceedings of the 61st Annual Meeting of the Association for Computational Linguistics (Volume 1: Long Papers). pp. 47--65. Association for Computational Linguistics, Toronto, Canada (2023), \url{https://arxiv.org/abs/1911.03109/}

\bibitem{pan2024umuteam}
Pan, R., Garc{\'{\i}}a-D{\'{\i}}az, J.A., Bernal-Beltr{\'a}n, T., Valencia-Garc{\'{\i}}a, R.: Umuteam at semeval-2024 task 6: Leveraging zero-shot learning for detecting hallucinations and related observable overgeneration mistakes. In: Proceedings of the 18th International Workshop on Semantic Evaluation (SemEval-2024). pp. 717--724 (2024)

\bibitem{Rykov2024smurfcat}
Rykov, E.S., Shishkina, Y., Petrushina, K., Titova, K., Petrakov, S., Panchenko, A.: Smurfcat at semeval-2024 task 6: Leveraging synthetic data for hallucination detection. In: Proceedings of the 18th International Workshop on Semantic Evaluation (SemEval-2024). pp. 869--880. Association for Computational Linguistics, Mexico City, Mexico (2024), \url{https://aclanthology.org/2024.semeval-1.94/}

\bibitem{MARiA2024}
Sanayei, R., Singh, A., Rezaei, M., Bethard, S.: Maria at semeval 2024 task-6: Hallucination detection through llms, mnli, and cosine similarity. In: Proceedings of the 18th International Workshop on Semantic Evaluation (SemEval-2024). pp. 1583--1588 (2024), \url{https://aclanthology.org/2024.semeval-1.225/}

\bibitem{OPDAI2024}
Wei, C., Chen, Z., Fang, S., He, J., Gao, M.: Opdai at semeval-2024 task 6: Small llms can accelerate hallucination detection with weakly supervised data. In: Proceedings of the 18th International Workshop on Semantic Evaluation (SemEval-2024). pp. 720--729 (2024), \url{https://aclanthology.org/2024.semeval-1.104/}

\end{thebibliography}
\end{document}